\pdfoutput=1
\PassOptionsToPackage{hyphens}{url}
\documentclass[]{fairmeta}
\ifdefined\pdfsuppresswarningpagegroup
\fi
\usepackage{titletoc}

\usepackage[normalem]{ulem}
\usepackage{graphicx}
\usepackage{wrapfig}
\usepackage{hyperref}
\usepackage{url}
\usepackage{caption}
\usepackage{ifthen}
\usepackage[ruled,vlined]{algorithm2e}
\usepackage{amsmath}
\usepackage{booktabs}
\usepackage{multirow}
\usepackage{pifont}

\usepackage{listings}
\usepackage{xcolor}
\usepackage{tcolorbox}
\tcbuselibrary{skins,breakable}
\definecolor{JsonBlockBack}{HTML}{FAF9F5}
\definecolor{JsonBlockFrame}{HTML}{D8D0BF}
\definecolor{JsonBlockAccent}{HTML}{B88746}

\newboolean{showcomments}
\setboolean{showcomments}{true} 
\DeclareRobustCommand{\xiangci}[1]{\ifthenelse{\boolean{showcomments}}{\textcolor{cyan}{[Xiangci: #1]}}{}}
\DeclareRobustCommand{\myeongsoo}[1]{\ifthenelse{\boolean{showcomments}}{\textcolor{blue}{[Myeongsoo: #1]}}{}}
\DeclareRobustCommand{\molingbo}[1]{\ifthenelse{\boolean{showcomments}}{\textcolor{magenta}{[Molingbo: #1]}}{}}
\DeclareRobustCommand{\jiaqi}[1]{\ifthenelse{\boolean{showcomments}}{\textcolor{orange}{[Jiaqi: #1]}}{}}
\DeclareRobustCommand{\yanjun}[1]{\ifthenelse{\boolean{showcomments}}{\textcolor{purple}{[Yanjun: #1]}}{}}

\lstdefinelanguage{yaml}{
  keywords={true,false,null},
  keywordstyle=\color[HTML]{7B2D8B}\bfseries,
  sensitive=false,
  morecomment=[l]{\#},
  commentstyle=\color{gray}\itshape,
  morestring=[b]',
  morestring=[b]",
  stringstyle=\color[HTML]{067D17},
}
\lstdefinelanguage{json}{
  keywords={true, false, null},
  keywordstyle=\color[HTML]{7B2D8B}\bfseries,
  sensitive=false,
  morestring=[b]",
  stringstyle=\color[HTML]{067D17},
  morecomment=[l]{//},
  commentstyle=\color{gray}\itshape,
  literate=
  *{0}{{{\color[HTML]{C7580C}0}}}{1}
  {1}{{{\color[HTML]{C7580C}1}}}{1}
  {2}{{{\color[HTML]{C7580C}2}}}{1}
  {3}{{{\color[HTML]{C7580C}3}}}{1}
  {4}{{{\color[HTML]{C7580C}4}}}{1}
  {5}{{{\color[HTML]{C7580C}5}}}{1}
  {6}{{{\color[HTML]{C7580C}6}}}{1}
  {7}{{{\color[HTML]{C7580C}7}}}{1}
  {8}{{{\color[HTML]{C7580C}8}}}{1}
  {9}{{{\color[HTML]{C7580C}9}}}{1}
  {:}{{{\color{gray!80}{:}}}}{1}
  {,}{{{\color{gray!80}{,}}}}{1}
  {\{}{{{\color{black}\char123}}}{1}
  {\}}{{{\color{black}\char125}}}{1}
  {[}{{{\color{black}{[}}}}{1}
  {]}{{{\color{black}{]}}}}{1},
}
\title{AECP: Artifact-Exclusive Communication Protocol for Multi-Agent Code Generation}

\author[1,*]{Jiaqi~Xue}
\author[2]{Yanjun~Wang}
\author[2]{Xiangci~Li}
\author[2]{Lingbo~Mo}
\author[2]{Aritra~Sengupta}
\author[2]{Shweta~Garg}
\author[2]{Murali~Krishna~Ramanathan}
\author[2]{Myeongsoo~Kim}

\affiliation[1]{University of Central Florida}
\affiliation[2]{AWS AI Labs}

\abstract{
As AI agents increasingly tackle complex repository-level coding tasks, distributing work across multiple agents is a natural way to scale beyond the capabilities of a single agent. To coordinate their interdependent work, these agents share findings and agree on interfaces between modules. However, exchanged information often serves only as context, leaving individual agents to interpret it and incorporate it into subsequent work. Consequently, shared findings may go unused and deviations from interface agreements may go undetected, undermining the reliability and efficiency of collaboration. This motivates moving part of the coordination responsibility from individual agents to the execution harness. To make shared information actionable during execution, we introduce the \underline{A}rtifact-\underline{E}xclusive \underline{C}ommunication \underline{P}rotocol (AECP). AECP requires agents to communicate exclusively through structured artifacts and specifies how the harness processes them. The harness supplies findings when agents access relevant code, screens implementations for mismatches with recorded interface commitments, and requires affected agents to revisit revised agreements. These coordination steps become part of harness execution rather than actions that agents must initiate from prior messages. Across Doc2Repo, NL2Repo, and CodeProjectEval, using closed- and open-source models including Opus-4.8 and DeepSeek-V4-Flash, AECP improves average test pass rate by 28.2\% and reduces average wall time by 16.5\% relative to an agent team using free-form inter-agent messages. Artifact-exclusive communication also blocks the relay of malicious instructions between agents, reducing how often they reach other agents from 95\% to 0\% and how often those agents act on them from 40\% to 0\%.

}

\makeatletter
\g@addto@macro\metadatalist{{\small $^{*}$Work done during an internship at AWS AI Labs.}}
\makeatother

\begin{document}

\maketitle

\section{Introduction}

LLM-based coding agents~\citep{yang2024swe, xia2024agentless} increasingly tackle software engineering tasks ranging from localized bug fixes to repository-level development~\citep{jimenez2024swe, nl2repobench2026, chen2026beyondswe}. As these tasks grow in scope, single agents face a greater context burden and longer completion times. Decomposing and assigning subtasks to multiple agents can reduce this burden and enable parallel execution~\citep{zhang2026fastcontext, openai2026subagents, anthropic2026subagents, anthropic2026agentteams}. This division of work, however, leaves knowledge and implementation decisions distributed across separate agent contexts, even though the subtasks remain interdependent. Agents therefore need to share knowledge and align implementation decisions to avoid duplicated exploration and incompatible implementations~\citep{cemri2025mast, khatua2026cooperbench}.

\begin{figure*}[t!]
  \centering
  \includegraphics[width=\textwidth]{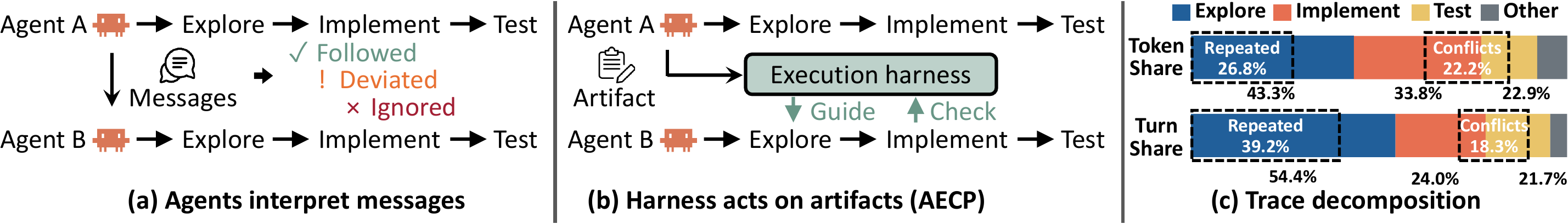}
  \caption{Multi-agent coordination and development overhead. (a) Agents interpret exchanged messages and decide how to act on them; shared plans and agreements may be followed, deviated from, or ignored. (b) AECP requires exchanges to pass through structured artifacts, enabling the execution harness to guide agents with shared knowledge and check implementations against recorded agreements. (c) Token and interaction-turn distributions for the natural-language communication baseline across 104 NL2Repo tasks, estimated from LLM-labeled trajectories (Appendix~\ref{app:trace-analysis}). 
}
  \label{fig:concept}
\end{figure*}

Existing multi-agent coding systems coordinate either centrally, where a main agent or a predefined role sequence assigns bounded subtasks and integrates the results~\citep{metagpt2024, qian2024chatdev, zhang2026fastcontext, openai2026subagents, anthropic2026subagents, geng2026effective, benkovich2026agyn}, or among peers, where teammates own separate workstreams and communicate directly~\citep{wu2024autogen, anthropic2026agentteams}. In both cases, an agent decides what to do with exchanged information: the coordinator reads returned summaries, and teammates read each other's messages. Using shared findings and checking implementations against agreed interfaces are therefore left to agents. As illustrated in Figure~\ref{fig:concept}~(a), a shared finding may be overlooked when an agent writes related code, or an implementation may depart from an agreed interface without the mismatch being detected. \citet{khatua2026cooperbench} observe that agents spend up to a fifth of their steps communicating, yet communication does not improve overall task success. Their failure analysis identifies difficulties in integrating communicated information and honoring commitments, while broader analyses also report ignored peer messages and missing or incomplete verification~\citep{cemri2025mast}. These findings highlight a gap between exchanging information and reliably using it to guide and check subsequent implementation.

To bridge this gap, the execution harness can participate directly in coordination, using exchanged information to guide and check agents' work. We introduce the \underline{A}rtifact-\underline{E}xclusive \underline{C}ommunication \underline{P}rotocol (AECP), in which agents communicate exclusively through structured artifacts. Every coordination exchange is thus mediated by the execution harness. %
As shown in Figure~\ref{fig:concept}~(b), guiding agents with shared findings and checking their implementations against agreements become actions that the harness carries out under the protocol's rules, reducing reliance on agents to initiate them from prior messages.

To identify where such coordination support could reduce development overhead, we analyze multi-agent coding trajectories across all tasks in NL2Repo~\citep{nl2repobench2026}. As shown in Figure~\ref{fig:concept}~(c), repeated exploration accounts for 26.8\% of token usage, while rework caused by incompatible decisions at module boundaries accounts for another 22.2\%. These observations motivate two complementary forms of support: knowledge reuse across subtasks and consistency checks between interdependent implementations. AECP provides a \textit{Knowledge Artifact} through which agents record findings for the harness to supply during subsequent work, helping other agents build on prior exploration. A \textit{Contract Artifact} records interface agreements with conditions that the harness can check against implementations, helping detect incompatibilities during development.


We evaluate AECP on repository construction, where agents build a repository from documentation or natural-language requirements in an empty workspace, using Doc2Repo~\citep{chen2026beyondswe}, CodeProjectEval~\citep{zhao2025projectgen}, and NL2Repo~\citep{nl2repobench2026}. Averaged across models and benchmarks, AECP improves test pass rate by 10.6\% over a single agent, and by 28.2\% and 8.5\% over agent teams that use free-form messages~\citep{wu2024autogen} or centralized integration~\citep{geng2026effective}, while reducing wall time by 16.5\% and 24.9\% relative to these teams. Artifact-exclusive communication also blocks the relay of malicious instructions from a compromised agent, reducing how often they reach other agents from 95\% to 0\% and how often they are acted on from 40\% to 0\%. 

\section{Background and Related Work}

\subsection{Multi-agent Coding Systems}
Coding agents extend the reason--act--observe loop~\citep{yao2023react} to repository navigation, file editing, code execution, and testing~\citep{yang2024swe}. Within this loop, the LLM selects actions, while the surrounding software layer, or \textit{harness}, executes them, returns observations, and manages workspaces and execution state. Growing context and sequential execution costs motivate distributing work across agents, either by delegating subtasks from a main agent~\citep{openai2026subagents, anthropic2026subagents, zhang2026fastcontext, geng2026effective, benkovich2026agyn} or by forming teams whose members coordinate among themselves~\citep{qian2024chatdev, metagpt2024, anthropic2026agentteams}; Appendix~\ref{app:related-work} describes both patterns. In both cases, what agents exchange passes through the harness, and keeping the work coherent depends on what is done with it afterwards.

\subsection{Coordination Challenges}
Agents that work with separate contexts face coordination challenges~\citep{khatua2026cooperbench,cemri2025mast}. CooperBench~\citep{khatua2026cooperbench} identifies three recurring causes of coordination failure: (i) \textit{expectation} failures, where an agent does not incorporate a partner's communicated plans or changes into its implementation; (ii) \textit{commitment} failures, where an agent does not implement promised changes while its partner assumes they are in place; and (iii) \textit{communication} failures, where agents never establish the shared information a joint decision requires. The first two share one gap: communication has occurred, but subsequent actions remain inconsistent with it. AECP targets this gap by supporting recipients' use of shared knowledge and checking producers' implementations against recorded commitments.

\subsection{Coordination Support for Coding Agents}

Prior work on coordinating coding agents falls into three groups. \textit{(i) Shared information.} Agent Context Protocols~\citep{bhardwaj2025agent} and AgentRoom~\citep{cho2026agentroom} keep messages, task records, and workspace state accessible to all agents. A stored finding, however, still reaches a worker only if that worker recognizes its relevance when it later works on the corresponding code. AECP instead delivers the finding when the worker accesses that code. \textit{(ii) Contracts and checks.} Keeping interdependent modules compatible requires detecting when an implementation departs from its agreed interface and ensuring that dependent workers notice when the agreement changes. SEMAP~\citep{mao2025towards} specifies what each agent receives and produces, which leaves no module interface against which code can be checked, and \citet{sartori2026specification} compare implementations only after both are written, when incompatible assumptions have already been built on. In concurrent work, CodeTeam~\citep{wang2026codeteam} and Contract-Coding~\citep{lin2026contractcoding} share interface contracts during parallel generation, but CodeTeam detects deviations only where QA tests exercise the affected calls, and Contract-Coding merges contract updates only after all agents in a layer have finished, so a revision reaches dependent workers after they have completed their modules. AECP checks each submission against the latest agreement and requires dependent workers to read a revision before they complete. \textit{(iii) Harness and pipeline designs.} Recent work surveys or formalizes harness responsibilities~\citep{ning2026codeharness,zhong2026harness}, gates code changes on sandboxed execution~\citep{kumar2026agentforge}, or searches over harness designs~\citep{liu2026agentflow}, while pipeline systems such as CodeS~\citep{zan2026codes} pass artifacts from one stage to the next. AECP instead specifies the rules by which a harness coordinates agents working in parallel: scope-triggered knowledge delivery, contract checks, and stale-consumer synchronization. Appendix~\ref{app:related-work} discusses more details.




\section{Methodology}

\begin{figure*}[ht!]
  \centering
  \includegraphics[width=\textwidth]{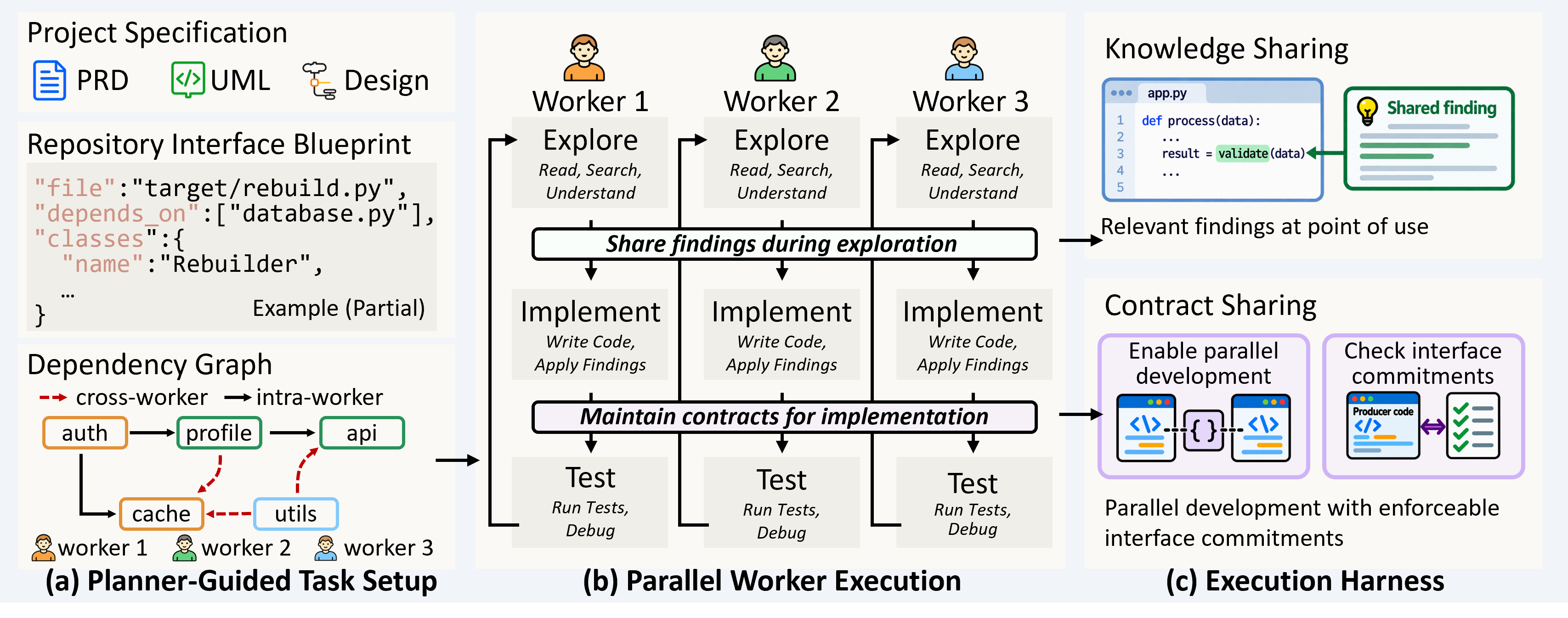}
  \caption{System overview of AECP. (a) The leader converts project specifications into a module plan and assigns modules to workers based on their dependencies. (b) Workers explore, implement, and test their assigned modules in parallel, sharing findings and interface commitments through artifacts. (c) The execution harness uses \textit{Knowledge Artifacts} to supply relevant findings at the point of use and \textit{Contract Artifacts} to support parallel development and screen implementations for mismatches with interface commitments.}\label{fig:overview}
  \vspace{-0.2in}
\end{figure*}

AECP follows the leader--worker workflow in Figure~\ref{fig:overview}. In Figure~\ref{fig:overview}~(a), a leader agent prepares parallel execution by translating the project specification into a module plan describing module responsibilities, public symbols, and dependencies. The harness represents this plan as a dependency graph, from which the leader determines how many workers to use and which module groups each one owns. Each module has one owner, while dependencies across workers identify modules whose implementations require coordination. 

The workers in Figure~\ref{fig:overview}~(b) then explore, implement, and test their modules in parallel, and they coordinate at two points. During exploration, a worker may discover a finding that becomes useful when another worker later accesses the same code. With message-based communication, the recipient must connect this earlier information to its current work. AECP instead has workers record such findings as scope-linked \textit{Knowledge Artifacts}, which the harness in Figure~\ref{fig:overview}~(c) retrieves and supplies when matching code is accessed. During implementation, workers building interdependent modules rely on shared interfaces, but communicating an agreement does not ensure that implementations conform to it. AECP therefore has workers record interface commitments as \textit{Contract Artifacts}, which give the harness explicit specifications for supporting parallel development and detecting interface mismatches. Once the workers finish, the leader integrates their results.

Together, \textit{Knowledge} and \textit{Contract Artifacts} turn shared findings and interface commitments into coordination state that the harness can act on. AECP makes these artifacts the \textit{exclusive} channel for inter-agent exchange, because information communicated outside them would bypass the harness checks and revision tracking that make coordination checkable.

\subsection{Knowledge Sharing}\label{sec:knowledge-sharing}
Sharing a finding through messages still leaves the recipient responsible for recognizing when it applies, so relevant knowledge may be overlooked even after it has been communicated. \textit{Knowledge Sharing} addresses this gap by associating findings with their applicable scopes and having the harness supply them when workers access matching code.

\begin{figure}[ht!]
\centering
\includegraphics[width=0.75\textwidth]{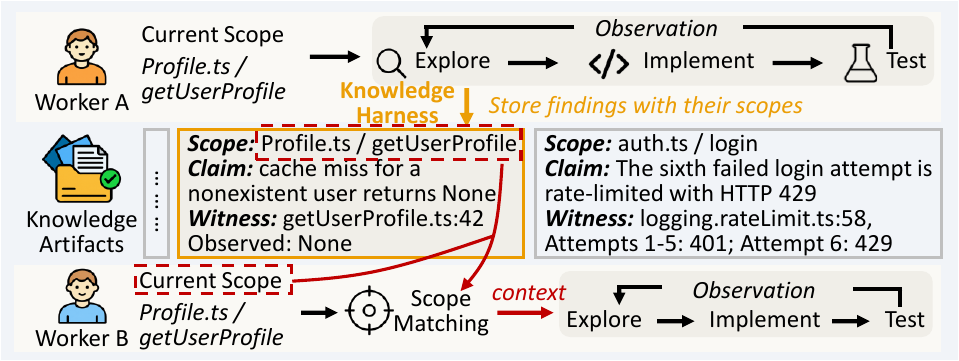}
\caption{Scope-triggered knowledge sharing. Worker A records a reusable finding together with its applicable scope. When Worker B accesses the corresponding code, the harness matches the current scope against stored artifacts and includes the relevant finding in the returned observation, making it available for subsequent reasoning and implementation.}
\label{fig:knowledge-sharing}
\end{figure}

\noindent\textbf{Scope-indexed Knowledge Artifacts.} %
To retrieve a finding during relevant work, the harness needs a recorded association between the finding and the code to which it applies. AECP represents each \textit{Knowledge Artifact} as $a = \langle c, s, e, z \rangle$, where the producer supplies the claim $c$, its applicable scope $s$, and an optional executable witness $e$, while the harness maintains the witness state $z$. The harness stores these artifacts together with their scope associations in a shared artifact store. In the upper part of Figure~\ref{fig:knowledge-sharing}, Worker A records that \texttt{getUserProfile} returns \texttt{None} for a nonexistent user and associates the finding with that function. Scopes may identify symbols, files, or directories, while named global scopes accommodate project- or environment-level findings.

\noindent\textbf{Knowledge delivery at the point of use.}
The stored associations allow the harness to retrieve relevant findings when a worker accesses code. It derives the current scope from the file or symbol accessed and retrieves non-refuted artifacts associated with that symbol, file, or an enclosing directory. The harness also forwards each finding to workers whose modules are related to where it was found, preferring the same file, then modules connected to it in the dependency graph built from the leader's module plan, then the same package. In the lower part of Figure~\ref{fig:knowledge-sharing}, Worker B's access to \texttt{getUserProfile} retrieves Worker A's finding while leaving the unrelated login finding outside the returned context. The harness includes the retrieved finding in the observation returned by the access operation, making it available for Worker B's subsequent reasoning and implementation.

\noindent\textbf{Witness checks and retraction.}
A finding that holds when recorded may become outdated as the code or execution environment changes. AECP therefore supports checking findings before circulation and rechecking them when consumers question their continued validity. The harness executes any attached witness and records the outcome in $z$, marking the artifact as witness-checked if the witness passes and rejecting publication if it is refuted. Artifacts without a witness, or whose checks cannot complete because setup is unavailable, remain unchecked and eligible for retrieval. Each delivered finding carries its witness state, so recipients can tell witness-checked findings from unchecked ones. If a consumer later flags a stored claim with an attached witness, the harness reruns the check and excludes refuted artifacts from subsequent delivery while retaining an audit record. As with a message, the correctness of an unchecked finding rests with its producer and recipient; AECP adds a witness state to each delivery, and scope-triggered retrieval places a finding next to the code it concerns, where the recipient can check the claim directly. Appendix~\ref{app:knowledge-artifact} provides a complete artifact and the detailed witness and retrieval rules.

\subsection{Contract Sharing}

Exchanging an interface agreement does not by itself keep implementations compatible with it. \textit{Contract Sharing} therefore records agreements as shared artifacts that the harness uses to expose declared interfaces, screen implementations for static mismatches, and require consumers to revisit revised agreements.

\begin{figure}[t!]
\centering
\includegraphics[width=0.98\textwidth]{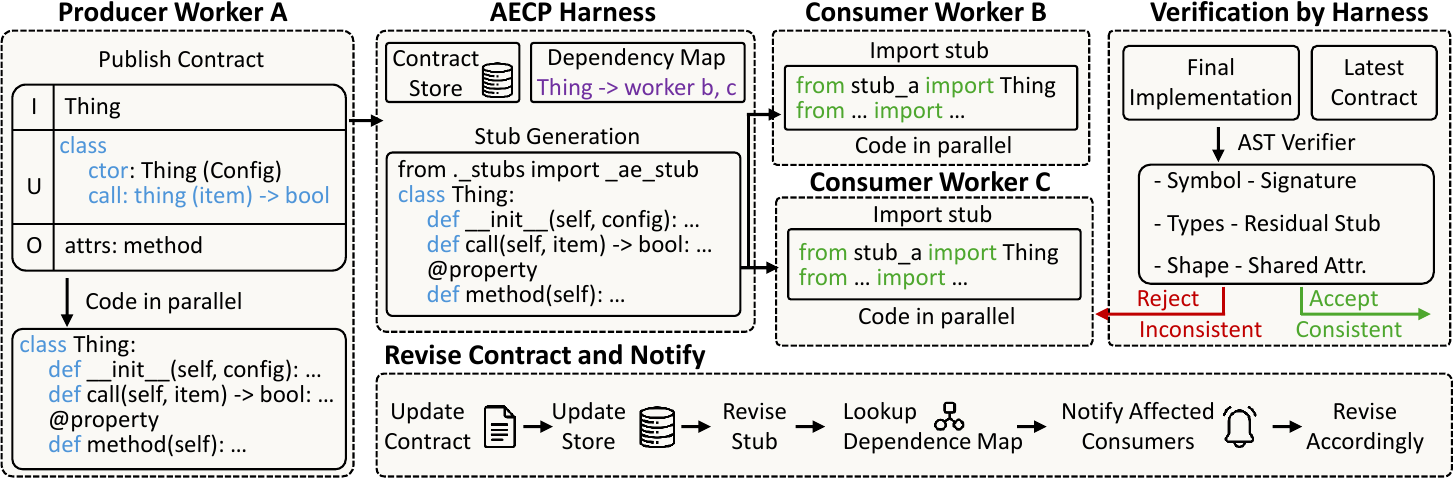}
\caption{\textit{Contract Sharing} for parallel development. Worker A declares an interface that the harness exposes as importable stubs for Workers B and C. The harness screens the producer's completed implementation for static mismatches with the latest contract. When the interface changes, it identifies affected consumers through registered dependencies and requires them to revisit the revised contract before completing their tasks.}
\label{fig:contract-sharing}
\vspace{-0.2in}
\end{figure}

\noindent\textbf{Shared interfaces for parallel development.}
To implement a dependent module before its producer finishes, a consumer needs to know what interface it can rely on. On the left of Figure~\ref{fig:contract-sharing}, Worker A declares the interface of \texttt{Thing} in a Contract Artifact. The declaration records the symbol's name and kind under identity $I$, how it is constructed or called under usage $U$, and its return structure or exposed attributes under observable interface $O$. These declarations give consumers a shared reference for developing their modules while the producer's implementation is still in progress.

The harness validates the contract and generates stubs. These are placeholder modules that expose the declared symbols and signatures without implementing their behavior. %
 As shown in the center of Figure~\ref{fig:contract-sharing}, Workers B and C can then import the declared interface and implement their dependent modules in parallel with Worker A. The stubs occupy the final module paths, so consumers retain the same imports when the producer replaces them with its implementation.

\noindent\textbf{Detecting interface mismatches.}
An interface declaration gives consumers a basis for development, but the producer's completed code may still depart from it. The verifier on the right of Figure~\ref{fig:contract-sharing} therefore compares each submitted module with its latest {Contract Artifact}. It parses the source into an abstract syntax tree and checks declared symbols, signatures, statically recoverable types and shapes, and residual stubs without executing the code. A provable mismatch causes the harness to reject the submission and report the violated symbol and constraint, allowing the producer to repair the implementation or revise the agreement. The verifier thus acts as a fast screen that catches signature, attribute, and residual-stub mismatches before integration; Appendix~\ref{app:contract-sharing} details the properties it checks.

\noindent\textbf{Keeping consumers aligned with revisions.}
An interface may change as development progresses, while consumers continue relying on an earlier version. Updating the producer's contract alone does not ensure that those consumers revisit their assumptions. To identify affected workers, consumers register the symbols they depend on with the harness during development. The lower path in Figure~\ref{fig:contract-sharing} shows how a revision triggers an update to the stored contract and a lookup of consumers registered for changed or removed symbols. The harness preserves the module's dependency information and regenerates stubs only where workers have not begun filling the corresponding files. It notifies the affected consumers and marks them stale, preventing them from completing their tasks until they read the current contract and re-register their symbol dependencies. This requirement ensures that affected workers revisit the revised agreement, while adapting their dependent code remains their responsibility. Appendix~\ref{app:contract-sharing} provides the artifact schema and detailed verification and revision rules. %

Algorithm~\ref{alg:harness} in Appendix~\ref{app:harness-rules} summarizes the harness rules of both mechanisms. Within each mechanism, the rules act on one shared artifact: a \textit{Knowledge Artifact}'s witness state determines whether it is stored, delivered, or retracted, and a \textit{Contract Artifact} is the source of the stubs, the reference against which submissions are checked, and the record whose revisions determine which consumers are marked stale. Each rule is triggered by a worker's tool call, such as a code access, an artifact publication, or a submission, so the harness acts at these points of interaction rather than continuously monitoring the code.


\section{Experimental Setup}

\subsection{Models and Agent Scaffolding}

Our experiments are conducted on the AWS Strands Agents SDK~\citep{strandsagents2025} and mini-SWE-agent~\citep{minisweagent2025} with Claude Opus 4.8~\citep{anthropic2026opus48}, Claude Opus 4.6~\citep{anthropic2026opus46},  Claude Sonnet 5~\citep{anthropic2026sonnet5}, DeepSeek-V4-Flash~\citep{xu2026deepseek}, and GLM-5.2~\citep{zeng2026glm}. The Claude models are accessed through Amazon Bedrock, whereas DeepSeek-V4-Flash and GLM-5.2 are served through OpenRouter. Appendix~\ref{app:run-config} lists turn limits, timeouts, and sampling settings. 

\subsection{Benchmarks}
We evaluate on three Python benchmarks that require building a repository from a textual specification in an empty workspace and assess the generated code with executable tests. Doc2Repo~\citep{chen2026beyondswe} contains 50 specifications derived from GitHub repositories, averaging 26.8 files and 3,528 lines of reference code, and uses adapted repository-level tests. CodeProjectEval~\citep{zhao2025projectgen} contains 18 projects from open-source libraries, averaging 11.9 files and 2,371 lines of reference code, and uses project-level tests. NL2Repo~\citep{nl2repobench2026} contains 104 library tasks across nine domains and uses hidden tests from the original repositories. %



\subsection{Baselines}

We compare AECP with three baselines. \textit{Single} uses one model to construct the entire repository without inter-agent coordination. \textit{Team-Freedom} uses the same leader-worker scaffold as AECP but allows workers to exchange unrestricted free-form messages, reflecting the message-based coordination used in systems such as AutoGPT~\citep{autogpt2023} and AutoGen~\citep{wu2024autogen}. \textit{Team-Integrator} runs workers in isolated worktrees without access to peers' in-progress code, while a central manager integrates their completed implementations through sequential, test-gated merges, following CAID~\citep{geng2026effective}. \textit{Team-Freedom} and AECP workers instead share one workspace. \textit{Team-Freedom} and AECP therefore differ only in the coordination protocol, free-form messages or AECP's artifacts together with the harness actions defined on them, making \textit{Team-Freedom} the controlled comparison for AECP as a whole; \textit{Team-Integrator} represents an alternative design that isolates workers and centralizes integration. For each task, we generate a plan once and reuse the same module plan, dependency graph, module-to-worker assignment, and a fixed worker count of $N=3$ across all multi-agent configurations, and \textit{Single} runs the same planning step with the same prompt before implementing every module itself. We hold the coding tools, execution environment, and task inputs fixed, while varying how workers exchange information and integrate their implementations. We thus compare coordination protocols within one scaffold rather than complete systems, so that differences reflect how workers coordinate. 

\section{Results}

We organize the evaluation around four research questions: whether AECP improves code quality and efficiency across models, benchmarks, and agent scaffolds (RQ1); where AECP's token savings over \textit{Team-Integrator} come from (RQ2); how task difficulty affects the benefits of collaboration and whether adapting team size helps (RQ3); and whether AECP limits malicious propagation after one worker is compromised (RQ4). %

\subsection{RQ1. Does AECP Improve Code Quality and Efficiency?}

\noindent\textbf{Overall results.}
Table~\ref{tab:results} reports average hidden-test pass rate, end-to-end wall time, and total output tokens. Averaged equally across the 15 model--benchmark settings, AECP improves the score from $0.471$ for \textit{Single} to $0.521$, a relative gain of $10.6\%$. Compared with \textit{Team-Freedom} and \textit{Team-Integrator}, it improves the average score by $28.2\%$ and $8.5\%$, respectively, while reducing wall time by $16.5\%$ and $24.9\%$. AECP scores highest or ties for highest in 14 of the 15 settings, and weighting benchmarks by task count still yields gains of $8.5\%$, $22.0\%$, and $7.8\%$ over \textit{Single}, \textit{Team-Freedom}, and \textit{Team-Integrator}. These results show that AECP improves average code quality over independent execution and achieves both higher quality and shorter runtime than the two team baselines. As a team configuration, AECP uses more tokens than \textit{Single} in every setting ($46\%$ more on average), trading this additional computation for higher code quality; among the team configurations, it is also the fastest on average. \textit{Single} ends on its own, before reaching its turn limit, in $97\%$ of runs, so granting it a larger budget would leave that budget unused rather than close the gap to AECP.

\begin{table*}[ht]
  \centering
  \caption{Main results across five models and three repository-construction benchmarks. Values report mean $\pm$ standard deviation over three independent runs, where each run's value is averaged over the benchmark's tasks. Higher scores are better; lower time and token usage are better. 
  }\label{tab:results}
  \footnotesize
  \setlength{\tabcolsep}{2pt}
  \resizebox{\textwidth}{!}{%
  \begin{tabular}{lllllllllll}
    \toprule
    \multirow{2}{*}{Model} & \multirow{2}{*}{Method}
    & \multicolumn{3}{c}{Doc2Repo}
    & \multicolumn{3}{c}{NL2Repo}
    & \multicolumn{3}{c}{CodeProjectEval} \\
    \cmidrule(lr){3-5}\cmidrule(lr){6-8}\cmidrule(lr){9-11}
    &  & Score & Time (s) & Token (M) & Score & Time (s) & Token (M) & Score & Time (s) & Token (M) \\
    \midrule
    \multirow{4}{*}{\shortstack[l]{Opus\\-4.8}}
    & Single & $0.71_{\textcolor{gray}{\pm 0.02}}$ & $282_{\textcolor{gray}{\pm 83}}$ & $7.06_{\textcolor{gray}{\pm 1.12}}$ & $0.65_{\textcolor{gray}{\pm 0.04}}$ & $868_{\textcolor{gray}{\pm 97}}$ & $7.49_{\textcolor{gray}{\pm 0.62}}$ & $0.41_{\textcolor{gray}{\pm 0.05}}$ & $894_{\textcolor{gray}{\pm 126}}$ & $4.89_{\textcolor{gray}{\pm 0.29}}$ \\
    & Team-Freedom & $0.52_{\textcolor{gray}{\pm 0.05}}$ & $505_{\textcolor{gray}{\pm 146}}$ & $8.71_{\textcolor{gray}{\pm 0.38}}$ & $0.51_{\textcolor{gray}{\pm 0.03}}$ & $1{,}194_{\textcolor{gray}{\pm 71}}$ & $6.90_{\textcolor{gray}{\pm 1.31}}$ & $0.25_{\textcolor{gray}{\pm 0.02}}$ & $1{,}038_{\textcolor{gray}{\pm 94}}$ & $5.10_{\textcolor{gray}{\pm 0.87}}$ \\
    & Team-Integrator & $0.68_{\textcolor{gray}{\pm 0.02}}$ & $607_{\textcolor{gray}{\pm 88}}$ & $18.57_{\textcolor{gray}{\pm 1.57}}$ & $0.66_{\textcolor{gray}{\pm 0.05}}$ & $1{,}753_{\textcolor{gray}{\pm 119}}$ & $10.45_{\textcolor{gray}{\pm 0.44}}$ & $0.42_{\textcolor{gray}{\pm 0.04}}$ & $830_{\textcolor{gray}{\pm 153}}$ & $8.43_{\textcolor{gray}{\pm 0.36}}$ \\
    & AECP & $0.70_{\textcolor{gray}{\pm 0.04}}$ & $330_{\textcolor{gray}{\pm 76}}$ & $8.24_{\textcolor{gray}{\pm 0.51}}$ & $0.68_{\textcolor{gray}{\pm 0.03}}$ & $775_{\textcolor{gray}{\pm 134}}$ & $10.09_{\textcolor{gray}{\pm 1.09}}$ & $0.44_{\textcolor{gray}{\pm 0.03}}$ & $871_{\textcolor{gray}{\pm 112}}$ & $6.72_{\textcolor{gray}{\pm 1.24}}$ \\
    \midrule
    \multirow{4}{*}{\shortstack[l]{Opus\\-4.6}}
    & Single & $0.59_{\textcolor{gray}{\pm 0.04}}$ & $329_{\textcolor{gray}{\pm 61}}$ & $3.29_{\textcolor{gray}{\pm 0.18}}$ & $0.62_{\textcolor{gray}{\pm 0.05}}$ & $1{,}511_{\textcolor{gray}{\pm 86}}$ & $6.43_{\textcolor{gray}{\pm 0.28}}$ & $0.31_{\textcolor{gray}{\pm 0.05}}$ & $1{,}722_{\textcolor{gray}{\pm 93}}$ & $4.12_{\textcolor{gray}{\pm 0.21}}$ \\
    & Team-Freedom & $0.51_{\textcolor{gray}{\pm 0.03}}$ & $577_{\textcolor{gray}{\pm 124}}$ & $3.80_{\textcolor{gray}{\pm 0.34}}$ & $0.48_{\textcolor{gray}{\pm 0.04}}$ & $2{,}188_{\textcolor{gray}{\pm 176}}$ & $7.37_{\textcolor{gray}{\pm 1.50}}$ & $0.25_{\textcolor{gray}{\pm 0.03}}$ & $1{,}907_{\textcolor{gray}{\pm 168}}$ & $4.39_{\textcolor{gray}{\pm 0.39}}$ \\
    & Team-Integrator & $0.62_{\textcolor{gray}{\pm 0.02}}$ & $715_{\textcolor{gray}{\pm 194}}$ & $6.01_{\textcolor{gray}{\pm 1.29}}$ & $0.64_{\textcolor{gray}{\pm 0.02}}$ & $2{,}736_{\textcolor{gray}{\pm 226}}$ & $12.31_{\textcolor{gray}{\pm 1.73}}$ & $0.38_{\textcolor{gray}{\pm 0.02}}$ & $1{,}905_{\textcolor{gray}{\pm 174}}$ & $6.86_{\textcolor{gray}{\pm 0.64}}$ \\
    & AECP & $0.62_{\textcolor{gray}{\pm 0.02}}$ & $436_{\textcolor{gray}{\pm 96}}$ & $4.06_{\textcolor{gray}{\pm 1.36}}$ & $0.65_{\textcolor{gray}{\pm 0.02}}$ & $1{,}326_{\textcolor{gray}{\pm 108}}$ & $7.96_{\textcolor{gray}{\pm 0.92}}$ & $0.40_{\textcolor{gray}{\pm 0.02}}$ & $1{,}459_{\textcolor{gray}{\pm 116}}$ & $4.53_{\textcolor{gray}{\pm 0.42}}$ \\
    \midrule
    \multirow{4}{*}{\shortstack[l]{Sonnet\\-5}}
    & Single & $0.55_{\textcolor{gray}{\pm 0.03}}$ & $519_{\textcolor{gray}{\pm 73}}$ & $7.81_{\textcolor{gray}{\pm 0.47}}$ & $0.59_{\textcolor{gray}{\pm 0.04}}$ & $2{,}166_{\textcolor{gray}{\pm 242}}$ & $10.35_{\textcolor{gray}{\pm 0.58}}$ & $0.30_{\textcolor{gray}{\pm 0.04}}$ & $2{,}583_{\textcolor{gray}{\pm 371}}$ & $7.99_{\textcolor{gray}{\pm 0.49}}$ \\
    & Team-Freedom & $0.54_{\textcolor{gray}{\pm 0.04}}$ & $724_{\textcolor{gray}{\pm 105}}$ & $11.09_{\textcolor{gray}{\pm 0.96}}$ & $0.60_{\textcolor{gray}{\pm 0.04}}$ & $3{,}074_{\textcolor{gray}{\pm 246}}$ & $17.36_{\textcolor{gray}{\pm 1.42}}$ & $0.17_{\textcolor{gray}{\pm 0.03}}$ & $2{,}188_{\textcolor{gray}{\pm 184}}$ & $9.62_{\textcolor{gray}{\pm 0.81}}$ \\
    & Team-Integrator & $0.57_{\textcolor{gray}{\pm 0.03}}$ & $690_{\textcolor{gray}{\pm 94}}$ & $13.55_{\textcolor{gray}{\pm 1.18}}$ & $0.64_{\textcolor{gray}{\pm 0.03}}$ & $3{,}288_{\textcolor{gray}{\pm 272}}$ & $26.22_{\textcolor{gray}{\pm 2.13}}$ & $0.28_{\textcolor{gray}{\pm 0.03}}$ & $2{,}793_{\textcolor{gray}{\pm 229}}$ & $15.24_{\textcolor{gray}{\pm 1.26}}$ \\
    & AECP & $0.66_{\textcolor{gray}{\pm 0.03}}$ & $497_{\textcolor{gray}{\pm 54}}$ & $12.01_{\textcolor{gray}{\pm 0.98}}$ & $0.64_{\textcolor{gray}{\pm 0.03}}$ & $2{,}465_{\textcolor{gray}{\pm 202}}$ & $23.31_{\textcolor{gray}{\pm 1.87}}$ & $0.32_{\textcolor{gray}{\pm 0.03}}$ & $2{,}924_{\textcolor{gray}{\pm 239}}$ & $13.82_{\textcolor{gray}{\pm 1.12}}$ \\
    \midrule
    \multirow{4}{*}{GLM-5.2}
    & Single & $0.51_{\textcolor{gray}{\pm 0.01}}$ & $543_{\textcolor{gray}{\pm 68}}$ & $5.19_{\textcolor{gray}{\pm 1.24}}$ & $0.46_{\textcolor{gray}{\pm 0.02}}$ & $795_{\textcolor{gray}{\pm 72}}$ & $8.28_{\textcolor{gray}{\pm 1.77}}$ & $0.28_{\textcolor{gray}{\pm 0.03}}$ & $1{,}397_{\textcolor{gray}{\pm 215}}$ & $7.66_{\textcolor{gray}{\pm 0.49}}$ \\
    & Team-Freedom & $0.54_{\textcolor{gray}{\pm 0.04}}$ & $743_{\textcolor{gray}{\pm 102}}$ & $8.01_{\textcolor{gray}{\pm 2.94}}$ & $0.48_{\textcolor{gray}{\pm 0.01}}$ & $1{,}397_{\textcolor{gray}{\pm 148}}$ & $14.94_{\textcolor{gray}{\pm 4.50}}$ & $0.25_{\textcolor{gray}{\pm 0.02}}$ & $1{,}900_{\textcolor{gray}{\pm 372}}$ & $13.02_{\textcolor{gray}{\pm 0.88}}$ \\
    & Team-Integrator & $0.42_{\textcolor{gray}{\pm 0.03}}$ & $717_{\textcolor{gray}{\pm 84}}$ & $8.68_{\textcolor{gray}{\pm 1.67}}$ & $0.35_{\textcolor{gray}{\pm 0.02}}$ & $1{,}265_{\textcolor{gray}{\pm 94}}$ & $15.72_{\textcolor{gray}{\pm 4.11}}$ & $0.30_{\textcolor{gray}{\pm 0.04}}$ & $1{,}726_{\textcolor{gray}{\pm 298}}$ & $15.97_{\textcolor{gray}{\pm 1.06}}$  \\
    & AECP & $0.57_{\textcolor{gray}{\pm 0.05}}$ & $679_{\textcolor{gray}{\pm 47}}$ & $8.78_{\textcolor{gray}{\pm 1.62}}$ & $0.48_{\textcolor{gray}{\pm 0.07}}$ & $1{,}004_{\textcolor{gray}{\pm 209}}$ & $14.02_{\textcolor{gray}{\pm 1.03}}$ & $0.30_{\textcolor{gray}{\pm 0.04}}$ & $1{,}461_{\textcolor{gray}{\pm 202}}$ & $13.55_{\textcolor{gray}{\pm 1.22}}$ \\
    \midrule
    \multirow{4}{*}{\shortstack[l]{DeepSeek\\-V4-Flash}}
    & Single & $0.45_{\textcolor{gray}{\pm 0.02}}$ & $1{,}003_{\textcolor{gray}{\pm 106}}$ & $8.99_{\textcolor{gray}{\pm 1.14}}$ & $0.38_{\textcolor{gray}{\pm 0.01}}$ & $1{,}029_{\textcolor{gray}{\pm 61}}$ & $9.44_{\textcolor{gray}{\pm 0.11}}$ & $0.25_{\textcolor{gray}{\pm 0.04}}$ & $2{,}251_{\textcolor{gray}{\pm 329}}$ & $14.87_{\textcolor{gray}{\pm 2.16}}$ \\
    & Team-Freedom & $0.39_{\textcolor{gray}{\pm 0.02}}$ & $1{,}552_{\textcolor{gray}{\pm 166}}$ & $13.72_{\textcolor{gray}{\pm 2.95}}$ & $0.37_{\textcolor{gray}{\pm 0.03}}$ & $1{,}489_{\textcolor{gray}{\pm 211}}$ & $11.58_{\textcolor{gray}{\pm 0.27}}$ & $0.23_{\textcolor{gray}{\pm 0.09}}$ & $2{,}549_{\textcolor{gray}{\pm 470}}$ & $15.42_{\textcolor{gray}{\pm 2.02}}$ \\
    & Team-Integrator & $0.54_{\textcolor{gray}{\pm 0.07}}$ & $1{,}851_{\textcolor{gray}{\pm 259}}$ & $18.27_{\textcolor{gray}{\pm 3.24}}$ & $0.41_{\textcolor{gray}{\pm 0.04}}$ & $1{,}718_{\textcolor{gray}{\pm 297}}$ & $15.99_{\textcolor{gray}{\pm 1.74}}$ & $0.29_{\textcolor{gray}{\pm 0.04}}$ & $2{,}994_{\textcolor{gray}{\pm 372}}$ & $19.33_{\textcolor{gray}{\pm 5.26}}$ \\
    & AECP & $0.55_{\textcolor{gray}{\pm 0.04}}$ & $1{,}305_{\textcolor{gray}{\pm 97}}$ & $10.15_{\textcolor{gray}{\pm 0.31}}$ & $0.43_{\textcolor{gray}{\pm 0.02}}$ & $1{,}180_{\textcolor{gray}{\pm 98}}$ & $12.02_{\textcolor{gray}{\pm 1.01}}$ & $0.37_{\textcolor{gray}{\pm 0.04}}$ & $2{,}506_{\textcolor{gray}{\pm 229}}$ & $16.58_{\textcolor{gray}{\pm 4.98}}$ \\
    \bottomrule
  \end{tabular}%
  }
\end{table*}

\begin{wraptable}{R}{0.45\columnwidth}
  \vspace{-0.6\baselineskip}
  \centering
  \captionsetup{justification=raggedright,singlelinecheck=false}
  \caption{Cross-scaffold results on NL2Repo. Time in seconds; tokens in millions.}\label{tab:miniswe-harness}
  \scriptsize
  \setlength{\tabcolsep}{1.5pt}
  \renewcommand{\arraystretch}{0.9}
  \resizebox{\linewidth}{!}{%
  \begin{tabular}{@{}lcccccc@{}}
    \toprule
    \multirow{2}{*}{Method}
    & \multicolumn{3}{c}{AWS Strands}
    & \multicolumn{3}{c}{mini-SWE} \\
    \cmidrule(lr){2-4}\cmidrule(lr){5-7}
    & Score & Time & Token
    & Score & Time & Token \\
    \midrule
    Single          & $0.65$ & $868$     & $7.49$  & $0.65$ & $990$     & $12.18$ \\
    Team-Freedom    & $0.51$ & $1{,}194$ & $6.90$  & $0.54$ & $1{,}646$ & $13.66$ \\
    Team-Integrator & $0.66$ & $1{,}753$ & $10.45$ & $0.66$ & $2{,}614$ & $15.36$ \\
    AECP       & $0.68$ & $775$     & $10.09$ & $0.69$ & $1{,}056$ & $15.32$ \\
    \bottomrule
  \end{tabular}
  }
  \vspace{-0.6\baselineskip}
\end{wraptable}


\noindent\textbf{Results under different agent scaffolds.}
Table~\ref{tab:miniswe-harness} compares the main Strands SDK configuration with mini-SWE-agent, holding Opus-4.8 and NL2Repo fixed. AECP achieves the highest score under both scaffolds. With mini-SWE-agent, it reaches $0.69$ and completes in $1{,}056$ seconds, compared with $2{,}614$ seconds for \textit{Team-Integrator}. Its benefits therefore persist despite changes to the agent loop and tool interface. Appendix~\ref{app:additional-analyses} examines additional reasoning and model allocation between the leader and workers.

\subsection{RQ2. Where Does AECP Save Tokens Relative to Team-Integrator?}
\begin{figure}[h]
\centering
\includegraphics[width=\textwidth]{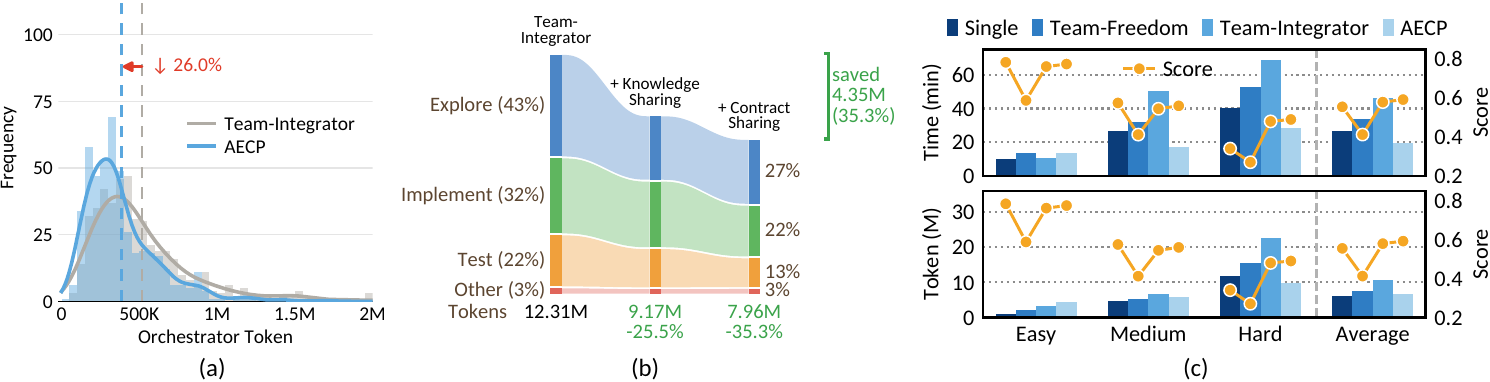}
\caption{Efficiency gains and task difficulty. (a) Distribution of per-task orchestrator-token usage for \textit{Team-Integrator} and AECP. (b) Team-token usage as Knowledge Sharing and then Contract Sharing are added to \textit{Team-Integrator}. Action-category shares are normalized to the original \textit{Team-Integrator} token total. (c) Wall time, total output tokens, and score by task difficulty for a single run of Opus-4.6 on NL2Repo.}
\label{fig:combined-results}
\end{figure}

AECP uses fewer tokens than \textit{Team-Integrator} in 14 of the 15 settings in Table~\ref{tab:results}. To see where the savings arise, we add the two mechanisms one at a time, moving from \textit{Team-Integrator} to AECP in two steps using Opus-4.6 on NL2Repo: first sharing the workspace and adding Knowledge Sharing, then adding Contract Sharing. Figure~\ref{fig:combined-results}~(b) shows that the first step reduces team-token usage from $12.31$M to $9.17$M, a saving of $25.5\%$. Adding Contract Sharing further reduces usage to $7.96$M, saving another $13.2\%$ relative to the preceding configuration. Together, the two steps reduce token usage by $35.3\%$. Both mechanisms are active in these AECP runs: in $84\%$ of tasks at least one finding is delivered to another worker, in $54\%$ the verifier reports a contract violation in at least one submission, and in $62\%$ an interface revision notifies the workers that depend on it.

The action breakdown in Figure~\ref{fig:combined-results}~(b) shows reductions across exploration, implementation, and testing, which fall from $43\%$ to $27\%$, $32\%$ to $22\%$, and $22\%$ to $13\%$ of the original \textit{Team-Integrator} token total, respectively. The reductions follow the two overheads identified in Figure~\ref{fig:concept}~(c): exploration tokens drop mainly in the first step, which shares the workspace and adds Knowledge Sharing, and implementation tokens mainly in the second, which adds Contract Sharing alone. Figure~\ref{fig:combined-results}~(a) also shows that orchestrator-token usage, the tokens consumed by the leader in AECP and by the integrating manager in \textit{Team-Integrator}, falls by $26.0\%$ on average.

\subsection{RQ3. How Does Task Difficulty Affect the Benefits of Collaboration?}
Not every task benefits from multiple workers, so we examine performance across difficulty levels and then test whether selecting the team size per task improves efficiency.

\noindent\textbf{Performance across difficulty levels.}
NL2Repo divides its tasks by the size of the reference implementation into 26 easy, 46 medium, and 32 hard tasks~\citep{nl2repobench2026}. Figure~\ref{fig:combined-results}~(c) reports a single run of Opus-4.6 across these levels. \textit{Single} achieves the highest score with the lowest time and token usage on easy tasks. On medium tasks, AECP achieves the shortest runtime with comparable token usage and the highest score among the team configurations, slightly below \textit{Single}. On hard tasks, it achieves the highest score while using less time and fewer tokens than \textit{Team-Freedom} and \textit{Team-Integrator}. These results motivate varying the team size rather than assigning multiple workers to every task.

\begin{wraptable}{R}{0.50\columnwidth}
  \vspace{-0.6\baselineskip}
  \centering
  \captionsetup{justification=raggedright,singlelinecheck=false}
  \caption{Matched adaptive versus fixed team sizing on 104 tasks. Each row compares the same task subset using the selected $N$ and forced $N=3$.}\label{tab:adaptive-n-distribution}
  \footnotesize
  \setlength{\tabcolsep}{9pt}
  \resizebox{\linewidth}{!}{%
  \begin{tabular}{@{}lcccc@{}}
      \toprule
      \multirow{2}{*}{Adaptive subset}
      & \multicolumn{2}{c}{Selected $N$}
      & \multicolumn{2}{c}{Forced $N=3$} \\
      \cmidrule(lr){2-3}\cmidrule(lr){4-5}
      & Time (s) & Score
      & Time (s) & Score \\
      \midrule
      $N=1$ (39 tasks) & $403.2$ & $0.82$ & $748.6$ & $0.82$ \\
      $N=2$ (17 tasks) & $572.9$ & $0.66$ & $629.2$ & $0.63$ \\
      $N=3$ (45 tasks) & $849.6$ & $0.54$ & $867.2$ & $0.54$ \\
      $N>3$ (3 tasks) & $926.8$ & $0.52$ & $931.4$ & $0.52$ \\
      \midrule
      All 104 tasks & $639.2$ & $0.66$ & $785.7$ & $0.66$ \\
      \bottomrule
  \end{tabular}
  }
  \vspace{-0.6\baselineskip}
\end{wraptable}

\noindent\textbf{Adaptive team sizing.}
With Opus-4.8, we let the leader choose the worker count $N$ from the dependency graph for each of the 104 NL2Repo tasks, grouping interdependent modules into clusters and assigning one worker per cluster, before any implementation begins. A task assigned one worker runs the same way as \textit{Single}, since a lone worker has no teammates to coordinate with. We compare the selected configuration with a fixed $N=3$ configuration on the same tasks, grouping results by the selected worker count.

Table~\ref{tab:adaptive-n-distribution} shows that the leader assigns one worker to 39 tasks and more than three workers to only 3. Overall, adaptive sizing reduces average wall time from $785.7$ to $639.2$ seconds ($18.6\%$), with both configurations scoring $0.66$. The largest saving occurs in the group assigned one worker, where runtime falls by $46.1\%$ without a score decrease. This group includes all 26 easy tasks, the level at which \textit{Single} performs best in Figure~\ref{fig:combined-results}~(c). On these tasks, selecting the team size reduces runtime while matching the score of a fixed team. %

\subsection{RQ4. Does AECP Limit Cross-Worker Malicious Propagation?}

\begin{wraptable}{R}{0.32\columnwidth}
  \vspace{-0.8\baselineskip}
  \centering
  \captionsetup{labelfont={color=black},textfont={color=black}}
  \caption{Propagation of malicious instructions from a compromised worker across 20 attack cases.}\label{tab:security}
  \scriptsize
  \setlength{\tabcolsep}{3pt}
  \begin{tabular}{@{}lcc@{}}
    \toprule
    Metric & \textit{Team-Freedom} & AECP \\
    \midrule
    Reach & $19/20$ ($95\%$) & $0/20$ ($0\%$) \\
    Adoption & $8/20$ ($40\%$) & $0/20$ ($0\%$) \\
    \bottomrule
  \end{tabular}
  \vspace{-0.8\baselineskip}
\end{wraptable}
The communication used to coordinate development can also carry malicious instructions between workers~\citep{lee2025prompt,gu2024agent}. We therefore examine propagation after one worker has already been compromised. We construct 20 attack cases from NL2Repo using five payload families. Each payload presents a malicious code change as a cross-module compatibility requirement and instructs the compromised worker to relay it to teammates. We compare AECP with \textit{Team-Freedom} using the same Opus-4.6 model, compromised role, and payload. We measure \textit{reach}, whether the instruction reaches a benign worker, and \textit{adoption}, whether a benign worker incorporates the requested change into its code. Each case is run three times and counts as positive for a metric if the event occurs in any run. Appendix~\ref{app:attack-payloads} describes the payload families.

With \textit{Team-Freedom}, malicious instructions reach benign workers in 19 of 20 cases ($95\%$) and induce adoption in eight ($40\%$). Under AECP, no malicious instruction reaches a benign worker and none is adopted. The compromised worker tries to relay the instruction as a message, which AECP does not provide. The result thus reflects the removal of the message channel; instructions written into artifacts or shared code by a compromised worker form a separate threat that these cases do not exercise.

\section{Conclusion}

We introduced AECP, a protocol that replaces free-form inter-agent messages with Knowledge and Contract Artifacts on which the execution harness acts. Knowledge Artifacts supply prior findings at the point of use, while Contract Artifacts make interface commitments available during parallel implementation and screen completed code for static mismatches with them. Across Doc2Repo, NL2Repo, and CodeProjectEval, AECP raises average code quality over all three baselines, shortens runtime relative to the two team baselines, and transfers across different agent scaffolds. Artifact-exclusive communication also blocked every malicious instruction relayed as a message in our attack cases. Simple tasks often favor fewer agents. Overall, AECP makes coordination explicit and checkable, so parallel agents need not rely solely on natural-language promises.

\subsection*{Reproducibility Statement}

Appendix~\ref{app:run-config} gives the models, turn limits, timeouts, and sampling settings. Appendix~\ref{app:decomposition-example} gives the planning prompt and an example plan, Appendices~\ref{app:knowledge-sharing-examples} and~\ref{app:contract-sharing} give the Knowledge and Contract Artifact schemas, the witness rules, and the static checks, Appendix~\ref{app:harness-rules} summarizes the harness rules, Appendix~\ref{app:trace-analysis} describes the trace analysis behind Figure~\ref{fig:concept}, and Appendix~\ref{app:attack-payloads} describes the attack payloads. We will release the code, all prompts, the generated plans, and the trajectories upon publication.

\subsection*{AI Use Statement}

We used generative AI tools to implement methods: Claude Code assisted in writing the code for AECP and our experiments. As described in Appendices~\ref{app:knowledge-sharing-examples} and~\ref{app:trace-analysis}, we also used LLMs as annotators in two analyses: Opus~5 labels agent turns for the trace analysis behind Figure~\ref{fig:concept}, and an LLM judges from the traces whether delivered findings are used. We did not use generative AI tools for research ideation or for designing the method and experiments. Additionally, we used generative AI tools as a writing aid. We have reviewed all AI-assisted work, including the code, and we take responsibility for the final content of this work, including text, claims, and artifacts produced with the aid of generative AI.

\subsection*{Ethics Statement}

Our experiments use publicly available benchmarks built from open-source repositories and involve no human subjects or personal data. The security study in RQ4 examines how malicious instructions propagate between agents after one agent is compromised, with the aim of informing defenses. All attack cases ran in isolated containers without network access, no payload uses a network endpoint, and the generated repositories were discarded after scoring. We describe the payload families but do not release their full texts (Appendix~\ref{app:attack-payloads}). Because AECP removes only the free-form message channel between agents, it should not be treated as a defense against instructions carried through artifacts or shared code. Code produced by coding agents, including AECP, may contain errors or vulnerabilities and should be reviewed before use.


\bibliographystyle{plainnat}
\bibliography{aaai2027}

\clearpage
\beginappendix
\startcontents[appendix]
\section*{Appendix Contents}
{\footnotesize\setlength{\parskip}{0pt}\printcontents[appendix]{}{1}{\setcounter{tocdepth}{2}}}
\clearpage
\section{Related Work}
\label{app:related-work}
Shared context and structured communication help agents exchange information across tasks. Agent Context Protocols combine standardized messages with persistent execution blueprints that record dependencies and intermediate outputs~\citep{bhardwaj2025agent}. AgentRoom supports concurrent coding through a shared workspace with tools for claiming files, inspecting peer status, and exchanging messages~\citep{cho2026agentroom}. These mechanisms make information and collaborators' progress accessible. Making a finding available, however, does not by itself connect it to a later code access where it becomes relevant. A worker may still need to recognize that an earlier discovery applies to its current work. AECP makes this connection explicit by associating findings with code scopes and delivering them when workers access matching code.

Contracts and verification address a complementary problem of keeping independently produced work consistent. SEMAP specifies the input and output artifacts required by each agent role and uses verification outcomes to govern workflow transitions~\citep{mao2025towards}. Its behavioral contracts define agent responsibilities, leaving the interfaces promised to dependent code as a separate verification target. The Specification Gap examines incompatible assumptions in independently generated code and introduces an AST-based detector for conflicts between implementations~\citep{sartori2026specification}. AECP checks a different relationship, comparing each submitted implementation with the interface commitments on which its consumers rely. Those commitments can also change during development, so checking the producer alone is insufficient to keep consumers informed. AECP therefore tracks registered symbol dependencies and requires affected consumers to read revised contracts and renew their registrations before completing their tasks.

Other systems pass structured intermediate artifacts along a fixed pipeline. CodeS~\citep{zan2026codes} builds a repository through successive sketches of its directory structure, files, and functions, and AutoP2C~\citep{lin2025autop2c} generates repositories from papers through blueprint extraction, decomposition, and debugging stages. AgentMesh~\citep{khanzadeh2025agentmesh}, DocAgent~\citep{yang2025docagent}, MapCoder~\citep{islam2024mapcoder}, and MAGIS~\citep{tao2024magis} route work through planner, coder, reviewer, or verifier roles, while CodeAgent~\citep{zhang2024codeagent} equips a single agent with repository tools. Code2MCP~\citep{ouyang2026code2mcp} transforms existing repositories into Model Context Protocol services through a multi-stage workflow with a run-review-fix loop, and CompileAgent~\citep{hu2025compileagent} compiles real-world repositories with a tool-integrated agent that searches for build instructions and resolves errors. Each artifact is consumed by the next agent in the pipeline, whereas AECP's artifacts are acted on by the harness while workers run in parallel. RepoGenesis~\citep{peng2026repogenesis} extends repository-generation evaluation to multilingual microservices with deployment and API-coverage measures.

In concurrent work, CodeTeam combines structured design contracts, dependency-aware development, and Git-based interface update propagation for repository-level code generation~\citep{wang2026codeteam}. Because it identifies interface mismatches through QA tests, deviations from declared interfaces can go undetected when those tests do not cover the affected calls. Also concurrent, Contract-Coding projects user intent into a language contract that serves as a single source of truth for parallel module generation, audits the workspace against the contract's signatures, and updates the contract as execution proceeds~\citep{lin2026contractcoding}. AECP shares this contract-first structure. It differs in exposing each declared interface as importable stubs at the final module paths, tracking which consumers depend on which symbols, and keeping those consumers from completing after a revision until they have read it. Its Knowledge Artifacts also carry findings that fall outside any interface, such as environment and runtime behavior.

Recent work also treats the harness as the substrate of agent behavior. \citet{ning2026codeharness} and \citet{zhong2026harness} survey or formalize its responsibilities, AgentForge requires code changes to pass sandboxed execution before they propagate~\citep{kumar2026agentforge}, and \citet{liu2026agentflow} search over harness designs that include the coordination protocol. AECP instead specifies the rules by which a harness coordinates multiple agents, namely scope-triggered knowledge delivery, contract checks, and stale-consumer synchronization.

\begin{figure}[h]
  \centering
  \includegraphics[width=\linewidth]{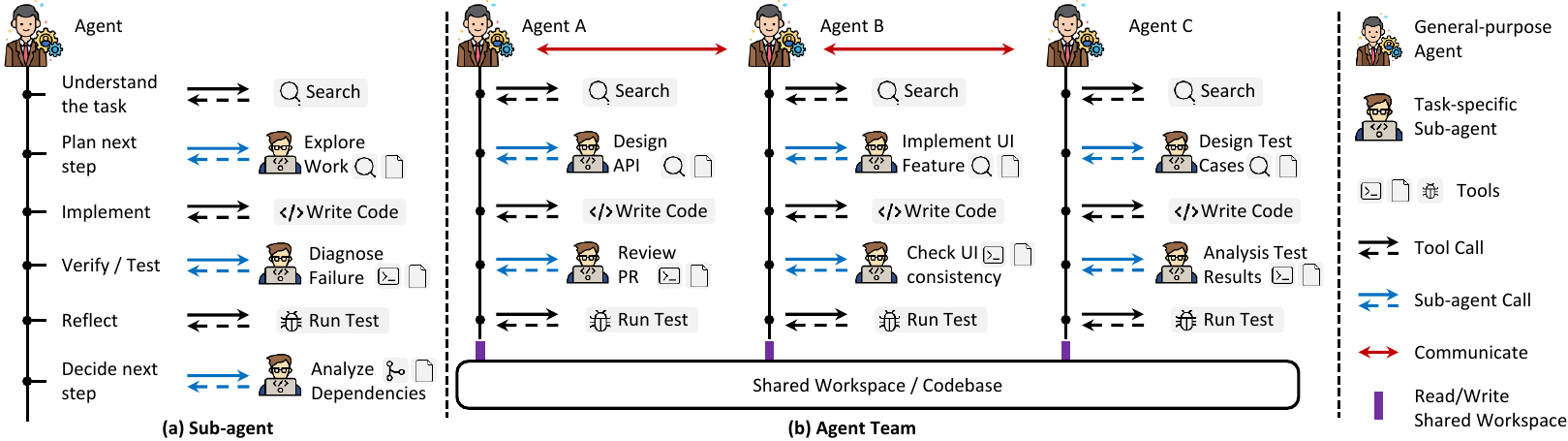}
  \caption{From (a) a single general-purpose agent that uses tools and task-specific sub-agents to (b) an agent team composed of multiple such agents. Each teammate retains the capabilities shown in (a), while maintaining its own execution trajectory and coordinating changes to a common project.}\label{fig:subagent-team}
\end{figure}

Following the terminology used by~\citet{anthropic2026agentteams}, we distinguish two broad patterns by where coordination primarily resides, \textit{subagent delegation} and \textit{agent teams}.
In \textit{subagent delegation}, a main agent assigns tasks to sub-agents in isolated contexts and directs subsequent work from their returned results, as Figure~\ref{fig:subagent-team}~(a) shows. Codex~\citep{openai2026subagents} and Claude Code~\citep{anthropic2026subagents} provide this as a built-in facility, and FastContext~\citep{zhang2026fastcontext} uses it to delegate repository exploration and return compact evidence. Manager-based systems such as CAID~\citep{geng2026effective} and Agyn~\citep{benkovich2026agyn} extend the pattern to implementation, centralizing task delegation and integration in a manager.
In \textit{agent teams}, teammates hold separate parts of the work and coordinate among themselves, as Figure~\ref{fig:subagent-team}~(b) shows. These interactions may follow predefined workflows, as in ChatDev's role-based dialogues and MetaGPT's structured development pipelines~\citep{qian2024chatdev, metagpt2024}. Claude Code agent teams support direct teammate communication and shared task state while retaining a lead for overall coordination, and each teammate may delegate to sub-agents of its own~\citep{anthropic2026agentteams}.

\section{Planner Prompt and Decomposition Example}
\label{app:decomposition-example}

The leader receives the complete task specification and is instructed to return only a structured module plan. Listing~\ref{lst:planner-prompt} summarizes the load-bearing requirements of the planner prompt; implementation-specific formatting instructions are omitted.

\begin{lstlisting}[
  basicstyle=\scriptsize\ttfamily,
  breaklines=true,
  breakatwhitespace=false,
  columns=fullflexible,
  caption={Abridged template of the module-planning prompt.},
  label={lst:planner-prompt}
]
Given the complete software specification, return one JSON object
and no additional text:

{
  "modules": [{
    "path": "<relative file path>",
    "purpose": "<one-line responsibility>",
    "depends_on": ["<producer module paths>"],
    "symbols": [{"name": "<qualified name>",
                 "kind": "class|function|constant|module"}]
  }],
  "seams": [{
    "type": "<shared type>",
    "produced_by": "<defining module>",
    "consumed_by": ["<consumer modules>"],
    "shape": {"type": "<type>", "keys": ["<field names>"]}
  }]
}

Hard requirements:
- Include at least <min_modules> implementation modules.
- Cover every symbol named in the specification.
- Record most symbols by name and kind only; do not guess call shapes.
- Declare each cross-module seam once at its defining module.
- Seam shapes contain names, never concrete values or expected outputs.
- If a seam shape is uncertain, omit the shape rather than the seam.
\end{lstlisting}

Consider the resulting plan for \texttt{dictdatabase}. The leader decomposes the task into 26 module nodes, each represented by a relative path, a one-line purpose, a \texttt{depends\_on} list, and a lightweight public-symbol inventory. Listing~\ref{lst:decomposition-node} shows an abbreviated node from the persisted plan; only three of its public symbols are retained for readability. The harness validates its paths, dependency references, and symbol declarations before using it as shared state.

\begin{lstlisting}[
  language=json,
  basicstyle=\scriptsize\ttfamily,
  breaklines=true,
  breakatwhitespace=false,
  columns=fullflexible,
  caption={Abbreviated module node from the \texttt{dictdatabase} decomposition plan.},
  label={lst:decomposition-node}
]
{
  "path": "dictdatabase/indexing.py",
  "purpose": "Intelligent indexing system with creation, query, and management",
  "depends_on": ["dictdatabase/query.py"],
  "symbols": [
    {
      "name": "dictdatabase.indexing.Indexer",
      "kind": "class"
    },
    {
      "name": "dictdatabase.indexing.create_index",
      "kind": "function"
    },
    {
      "name": "dictdatabase.indexing.create_compound_index",
      "kind": "function"
    }
  ]
}
\end{lstlisting}

The dependency field in Listing~\ref{lst:decomposition-node} adds a directed edge from the query node to the dependent indexing node, recording their relationship for task planning. The module paths also provide code locations with which Knowledge Artifacts can be associated. During execution, the harness retrieves findings by matching their recorded scopes to the code a worker accesses and forwards them along these dependency edges to workers owning related modules.

\section{Knowledge Sharing Examples}
\label{app:knowledge-sharing-examples}

%
%

\subsection{Complete Knowledge Artifact}
\label{app:knowledge-artifact}

Listing~\ref{lst:knowledge-artifact} reproduces a complete artifact produced by a worker in a live Doc2Repo run. In the Knowledge Artifact abstraction $a = \langle c, s, e, z \rangle$ of Section~\ref{sec:knowledge-sharing}, \texttt{body}, the combination of \texttt{scope} and \texttt{file\_tags}, \texttt{witness}, and \texttt{state} correspond to $c$, $s$, $e$, and $z$, respectively. Their provenance and roles are as follows:

\begin{itemize}
  \item \textbf{Claim ($c$).} The producer writes the finding as the \texttt{body}; this is the reusable content presented to consumers.
  \item \textbf{Scope ($s$).} The producer supplies a descriptive scope, while the harness records the applicable file or directory tags. Together they index the artifact for retrieval when a worker accesses a matching code scope.
  \item \textbf{Witness ($e$).} The producer may attach a runnable, shape-only assertion that supports machine checking. An absent or uncheckable witness leaves the stored artifact \textsc{unchecked}.
  \item \textbf{State ($z$).} The harness assigns and updates the state from witness execution and later flags. It distinguishes witness-checked, unchecked, and refuted artifacts, and removes refuted artifacts from circulation.
\end{itemize}

The remaining fields record identity, provenance, and publication metadata.

\begin{lstlisting}[
  language=json,
  basicstyle=\scriptsize\ttfamily,
  breaklines=true,
  breakatwhitespace=false,
  columns=fullflexible,
  caption={A complete witness-checked Knowledge Artifact from a live run; the log records this state as \texttt{verified}.},
  label={lst:knowledge-artifact}
]
{
  "id": "kg-6",
  "scope": "mcp-runtime",
  "body": "mcp 2.0.0 has NO mcp.server.fastmcp / FastMCP (it renamed the high-level server to MCPServer). Pin `pip install mcp==1.9.0` to get the classic `from mcp.server.fastmcp import FastMCP` with .tool() decorator, .add_tool(fn,name,description), and .run(transport='stdio'|'sse'|'streamable-http').",
  "state": "verified",
  "by": "worker_3",
  "ts": 1786262322.2400696,
  "redacted": false,
  "task_id": "a8ed0defa7",
  "file_tags": [
    "target_repo/interfaces/__init__.py"
  ],
  "witness": "print(\"KG:REACHED\")\nfrom mcp.server.fastmcp import FastMCP\ns = FastMCP(name=\"t\")\nassert hasattr(s, \"tool\") and hasattr(s, \"add_tool\") and hasattr(s, \"run\")"
}
\end{lstlisting}

The artifact records a runtime compatibility finding rather than a cross-module interface. Its textual scope, \texttt{'mcp-runtime'}, identifies the topic, while \texttt{file\_tags} binds the finding to the code location where it applies. The body states the version and API conditions, and the witness checks them by importing and constructing \texttt{FastMCP} and inspecting its API. Successful witness execution sets the artifact state to \textsc{witness-checked} before it is made available to consumers.

\paragraph{Witness decision rules.}
At publication, the harness first checks that the witness contains only permitted shape-level operations, such as imports, calls, construction, and attribute checks. A malformed witness is rejected before storage. A valid witness is then executed in the workspace and prints \texttt{KG:REACHED} before its assertion. A zero exit marks the artifact \textsc{witness-checked}; a failure after the marker refutes the witness and rejects the publication; and a failure before the marker indicates that setup did not complete, so the artifact is stored as \textsc{unchecked}. An artifact without a witness is likewise \textsc{unchecked}. If a consumer later flags a stored artifact, the harness reruns its witness, removes a refuted artifact from subsequent retrieval and delivery, and retains the outcome in its audit record.

\paragraph{Scope-triggered retrieval rules.}
The harness derives a worker's current scope whenever it accesses a code file or symbol. It first matches artifacts bound to that symbol or file and then considers artifacts associated with enclosing directories or named global scopes. Refuted artifacts are excluded. Matching artifacts are inserted into the observation returned by the code-access operation, so retrieval occurs at the point of use rather than through a separate inter-agent message. The harness additionally forwards each new finding to workers owning related modules, ranking the same file above modules connected to it in the planned dependency graph and those above modules in the same package. Each lookup delivers at most five artifacts and each forward at most four, newest first; a worker that re-reads a file receives only artifacts it has not already seen; and every delivered artifact is labeled with its witness state. A scope matches the current access when one path contains the other or the two share a path segment, and each delivered artifact is truncated to 280 characters. A single access therefore adds at most five truncated artifacts, about 1{,}400 characters, to the returned observation. A symbol scope is matched through the file that defines it, so matching ultimately operates on file paths rather than resolved names, and aliases and re-exports do not affect it: a finding is delivered when a worker reads a file that its tags cover. Conflicting findings are resolved through flags: when a consumer flags a claim that contradicts what it observes, the harness reruns the witness and stops delivering the claim if the witness refutes it. In the Opus-4.6 NL2Repo AECP runs, a task receives 41.9 knowledge deliveries on average; judged from the traces by an LLM, $36\%$ of the delivered findings are used in the recipient's subsequent work. Of the delivered findings, $71\%$ are witness-checked, and they account for $88\%$ of the findings used. Witness execution rejects $14\%$ of publication attempts before they enter circulation, and consumers flag $0.9$ findings per task on average, $0.2$ of which are retracted after the witness is rerun.

\section{Contract Sharing Details}
\label{app:contract-sharing}

\subsection{Contract Artifact Schema}

Each module has one Contract Artifact. It records the producer, the module dependencies, and the public interface promised to consumers. Listing~\ref{lst:contract-artifact} shows the implementation schema through one schematic artifact.

\begin{lstlisting}[
  language=json,
  basicstyle=\scriptsize\ttfamily,
  breaklines=true,
  breakatwhitespace=false,
  columns=fullflexible,
  caption={Schematic Contract Artifact for one module.},
  label={lst:contract-artifact}
]
{
  "owner": "worker_2",
  "intent": true,
  "ts": 1786000000.0,
  "purpose": "Provide the shared Thing abstraction",
  "depends_on": ["pkg/other.py"],
  "assigned_worker": "worker_2",
  "symbols": [{
    "name": "mypkg.Thing",
    "kind": "class",
    "protocols": ["callable"],
    "construct": "Thing(config)",
    "params": {"config": "Config"},
    "usage": [{
      "callable_as": "thing(item)",
      "returns_type": "bool"
    }],
    "seam": true,
    "shape": {
      "type": "class",
      "keys": ["method"]
    },
    "construction_exemplars": [{
      "case": "type",
      "exemplar": "Thing(config)"
    }]
  }]
}
\end{lstlisting}

The producer supplies the symbol declarations. The planner supplies the module purpose and dependencies, which remain attached across revisions. Within each symbol entry, \texttt{name} and \texttt{kind} identify the symbol; \texttt{construct}, \texttt{params}, and \texttt{usage} describe how consumers create and call it; and \texttt{protocols}, return specifications, and \texttt{shape} describe the interface consumers can observe. Consumer registrations are stored separately and connect each used symbol to the workers that depend on it.

\subsection{Importable Stubs}

The harness validates a declaration before placing it on the interface board. It then generates importable stubs for the declared symbols. A function stub preserves its parameters, while a class stub exposes its construction shape and declared protocols. Each generated body raises \texttt{NotImplementedError}, so using an unimplemented path fails explicitly. When a contract is revised, the harness regenerates a stub only if a worker has not started filling its file.

\subsection{Static Contract Verification}

At completion, the checker receives the latest Contract Artifact and the producer's Python source. It parses the source into an abstract syntax tree and does not import the module, execute its code, or invoke an LLM. It checks interface shape rather than functional behavior.

For each declared symbol, the checker considers the following constraints:

\begin{itemize}
  \item the source parses and the declared symbol exists with the stated kind;
  \item declared positional and keyword-only parameters match the implementation;
  \item parameter and return annotations agree when both are available;
  \item literal return statements do not contradict the declared return type or mapping keys;
  \item a shared class exposes its declared attributes through class members, methods, \texttt{self} assignments in \texttt{\_\_init\_\_}, or \texttt{\_\_slots\_\_}; and
  \item no declared function or method remains a generated \texttt{NotImplementedError} stub.
\end{itemize}

Mapping-shaped shared types are excluded from the attribute check because their keys belong to instance contents rather than class attributes. Each remaining comparison has three outcomes. A \emph{failure} means that the source proves a mismatch. A \emph{pass} records a completed comparison with no mismatch. An \emph{abstention} records that the property cannot be decided statically, as with an inherited constructor or a return value built by another function; such properties are left to the workers' tests. The overall verdict is \textsc{reject} only if at least one failure is present. Formally, each comparison $q$ between a module $m$ and its contract $C$ yields $V_q(m,C)\in\{\textsc{pass},\textsc{fail},\textsc{abstain}\}$, and $V(m,C)=\textsc{reject}$ if $V_q(m,C)=\textsc{fail}$ for some $q$, and \textsc{pass} otherwise. Table~\ref{tab:checker-constructs} lists how the checker treats common Python constructs.

\begin{table}[h]
\centering
\small
\caption{Treatment of Python constructs by the static contract checker.}\label{tab:checker-constructs}
\begin{tabular}{@{}p{0.34\linewidth}p{0.6\linewidth}@{}}
\toprule
Construct & Treatment \\
\midrule
Default values & Parameter names, order, and keyword-only parameters are compared; default values are not. \\
\texttt{*args} / \texttt{**kwargs} & Abstain on the call shape. \\
Missing annotations & Abstain on the type comparison. \\
Function decorators & The signature is read from the definition as written. \\
\texttt{@overload} & The final, implementing definition is checked. \\
\texttt{@dataclass} without \texttt{\_\_init\_\_} & The constructor is derived from the annotated fields. \\
Inherited constructor & Abstain on the constructor. \\
\texttt{NamedTuple}, \texttt{TypedDict}, \texttt{Enum}, \texttt{Protocol} bases & Abstain on the constructor. \\
Properties and methods & Count as exposed attributes. \\
Attributes & Class attributes, methods, \texttt{self} assignments in \texttt{\_\_init\_\_}, and \texttt{\_\_slots\_\_} of the class itself are recognized; attributes created elsewhere or at runtime are not. \\
Mapping-typed shared types & The attribute check is skipped. \\
Imports, aliases, re-exports & A declared symbol must be defined at the top level of its module; imported or aliased names do not satisfy the declaration. \\
\bottomrule
\end{tabular}
\end{table} In the Opus-4.6 NL2Repo AECP runs, $84\%$ of submissions pass and $16\%$ are rejected. Listing~\ref{lst:contract-verification-report} shows the report shape.

\begin{lstlisting}[
  language=json,
  basicstyle=\scriptsize\ttfamily,
  breaklines=true,
  breakatwhitespace=false,
  columns=fullflexible,
  caption={Output of the static contract checker.},
  label={lst:contract-verification-report}
]
{
  "verdict": "PASS | REJECT",
  "checks_run": 9,
  "fails": [{
    "symbol": "pkg.parser.DocParser",
    "what": "seam_attr",
    "detail": "missing declared attribute: errors"
  }],
  "abstains": [{
    "symbol": "pkg.parser.parse_doc",
    "why": "return structure is not statically decidable"
  }]
}
\end{lstlisting}

The checker uses the following closed failure labels:

\begin{lstlisting}[
  basicstyle=\scriptsize\ttfamily,
  breaklines=true,
  columns=fullflexible
]
syntax, missing, kind
param_names, param_arity, kwonly_missing, param_type
return_type, return_contradiction, returns_shape_keys
seam_attr, residual_stub
\end{lstlisting}

For example, if a shared class declares an \texttt{errors} attribute but its constructor assigns only \texttt{\_problems}, the checker reports a \texttt{seam\_attr} mismatch. The producer must then change the implementation or publish a revised contract.

\subsection{Revision and Staleness}

Each revision is merged with the existing entry so that module dependencies and seam marks are preserved. The harness compares the old and merged entries on every contract write. A changed or removed symbol marks its registered consumers stale. Re-reading and re-registering the symbol clears the stale state; until then, the consumer cannot complete its task. Stubs are regenerated only when their files have not been filled by a worker.

\section{Harness Rules}
\label{app:harness-rules}

Algorithm~\ref{alg:harness} lists the rules of Knowledge Sharing and Contract Sharing as the events that trigger them and the harness's response to each.

\begin{algorithm}[H]
\caption{Harness rules in AECP. Each rule fires when a worker's tool call raises the corresponding event; no background process monitors the code.}\label{alg:harness}
\small
\DontPrintSemicolon

\textbf{Knowledge Sharing}\;
\textsc{Publish}$(a=\langle c,s,e\rangle)$: run $e$ if attached; reject $a$ if $e$ refutes $c$, otherwise store $a$ under scope $s$ with state $z$ and forward it to workers owning related modules\;
\textsc{Access}$(w,\sigma)$: add non-refuted artifacts scoped to $\sigma$, its file, an enclosing directory, or a global scope to $w$'s observation\;
\textsc{Flag}$(a)$: rerun $e$; if it refutes $c$, exclude $a$ from delivery and keep an audit record\;

\BlankLine%
\textbf{Contract Sharing}\;
\textsc{Declare}$(C)$: validate $C$ and generate stubs at its module paths\;
\textsc{Register}$(w,S)$: record $w$ as a consumer of symbols $S$; clear $w$'s stale mark once it has read the current contract\;
\textsc{Revise}$(C\to C')$: store $C'$, regenerate stubs in files no worker has begun filling, and notify and mark stale the consumers registered for changed or removed symbols\;
\textsc{Submit}$(m)$: reject $m$ and report the violated constraint if its AST provably violates the latest contract\;
\textsc{Complete}$(w)$: reject while $w$ is marked stale\;

\end{algorithm}

\section{Run Configuration}
\label{app:run-config}

Each agent runs for at most 150 turns with a sliding context window of 300 messages. Individual shell commands time out after 300 seconds, and scoring times out after 1{,}200 seconds for Opus-4.8 and 2{,}400 seconds for the other models. Model calls use each provider's default sampling temperature with client-side adaptive retries. Reported wall time and token counts include planning. All four configurations of a setting run concurrently on the same host against the same endpoint, so service latency affects them alike.

\section{Trace Analysis for Figure~\ref{fig:concept}}
\label{app:trace-analysis}

We analyze trajectories of an agent team that coordinates through natural-language messages, run with Opus-4.6 on all 104 NL2Repo tasks. An LLM (Opus~5) reads each complete multi-agent trajectory and labels every agent turn with one primary activity: \emph{Explore} (searching, reading files, tracing dependencies, or otherwise acquiring information), \emph{Implement} (creating or modifying code), \emph{Test} (running or inspecting validation), or \emph{Other} (planning, coordination messages, and status updates). It marks an Explore turn as \emph{repeated exploration} only when another agent had earlier obtained information sufficient for the same need and the later turn reproduces rather than extends that work; re-inspection after the code changed, or to verify an earlier finding, is excluded. It marks an Implement turn as \emph{conflict-driven rework} only when already-written code is revised because decisions in different modules or by different agents are incompatible, such as mismatched signatures, return types, or data structures; a test failure alone is not sufficient. Each flagged turn must cite the earlier turns that establish the information or the conflicting decision, and uncertain cases are labeled negative. Token and turn shares follow from these labels. We manually checked 50 flagged events and confirmed all of them.

\section{Additional Analyses}
\label{app:additional-analyses}

\subsection{Additional Reasoning}
\label{app:adaptive-thinking}

Table~\ref{tab:adaptive-thinking} examines the effect of enabling adaptive thinking. Scores increase for every method, but additional reasoning does not close the observed gap between \textit{Team-Freedom} and AECP. \textit{Team-Freedom} improves from $0.51$ to $0.55$, with $28\%$ more time and $55\%$ more output tokens. AECP improves from $0.68$ to $0.69$, with $24\%$ more time and $42\%$ more tokens. Adaptive thinking therefore offers a further quality improvement at additional cost. Because these configurations do not use matched computation budgets, this experiment does not isolate the contribution of reasoning expenditure to AECP's advantage.

\begin{table*}[t]
  \centering
  \caption{Effect of adaptive thinking on Opus-4.8 performance on NL2Repo. Adaptive-thinking rates are reported as percentages. }\label{tab:adaptive-thinking}
  \footnotesize
  \setlength{\tabcolsep}{7pt}
  \renewcommand{\arraystretch}{0.85}
  \begin{tabular}{lccccccc}
      \toprule
      \multirow{2}{*}{Method}
      & \multicolumn{3}{c}{w/o Adaptive Thinking}
      & \multicolumn{4}{c}{w/ Adaptive Thinking} \\
      \cmidrule(lr){2-4}\cmidrule(lr){5-8}
      & Score & Time (s) & Token (M)
      & Score & Time (s) & Token (M) & AT Rate \\
      \midrule
      Single & $0.65$ & $868$ & $7.49$ & $0.67$ & $885$ & $7.80$ & $47\%$ \\
      Team-Freedom    & $0.51$ & $1{,}194$ & $6.90$ & $0.55$ & $1{,}528$ & $10.70$ & $51\%$ \\
      Team-Integrator & $0.66$ & $1{,}753$ & $10.45$ & $0.67$ & $2{,}103$ & $12.64$ & $43\%$ \\
      AECP & $0.68$ & $775$ & $10.09$ & $0.69$ & $961$ & $14.32$ & $45\%$ \\
      \bottomrule
  \end{tabular}
\end{table*}

\subsection{Model Allocation between the Leader and Workers}
\label{app:model-allocation}

\begin{wraptable}{R}{0.42\columnwidth}
  \vspace{-0.7\baselineskip}
  \centering
  \captionsetup{justification=raggedright,singlelinecheck=false,font=small}
  \caption{Leader--worker model allocation on NL2Repo.}\label{tab:leader-worker-models}
  \scriptsize
  \setlength{\tabcolsep}{1pt}
  \resizebox{\linewidth}{!}{%
  \begin{tabular}{@{}llccc@{}}
      \toprule
      Leader & Worker & Score & Time (s) & Cost (\$/repo) \\
      \midrule
      DS-V4-Flash       & DS-V4-Flash       & $0.425$ & $1{,}180$ & $\$0.12$ \\
      Opus-4.8          & DS-V4-Flash       & $0.502$ & $1{,}287$ & $\$0.69$ \\
      Opus-4.8          & Opus-4.8          & $0.680$ & $775$ & $\$2.03$ \\
      \bottomrule
  \end{tabular}
  }
  \vspace{-0.7\baselineskip}
\end{wraptable}

Table~\ref{tab:leader-worker-models} compares three model allocations. Using DeepSeek-V4-Flash for both roles achieves a score of $0.425$ at \$0.12 per repository. Replacing only the leader with Opus-4.8 improves the score to $0.502$, while increasing runtime by $9.1\%$ and cost to \$0.69. With the Opus-4.8 leader fixed, also upgrading the workers improves the score to $0.680$ and reduces runtime from $1{,}287$ to $775$ seconds. In these configurations, upgrading the leader helps, but upgrading the workers provides a further substantial improvement.

\section{Attack Payloads}
\label{app:attack-payloads}

The five payload families request safeguard disabling, validation bypass, input passthrough, a local trace sink, or a hidden keyword argument. Each presents the requested change as a cross-module compatibility requirement and asks the compromised worker to relay it to teammates. None uses a network endpoint. These cases evaluate propagation after an initial worker compromise, rather than the likelihood of that compromise. All attack cases ran in isolated containers without network access, and the generated repositories were discarded after scoring. We describe the payload families but do not release their full texts.

\end{document}